\documentclass{article}
\usepackage{spconf,amsmath,graphicx}

\title{Enhancing task-oriented dialogues with chitchat: a comparative study based on lexical diversity and divergence}
\name{Armand Stricker, Patrick Paroubek}
\address{Université Paris-Saclay, CNRS, \\
Laboratoire Interdisciplinaire des Sciences du Numérique, \\
91400, Orsay, France \\}
\copyrightnotice{979-8-3503-0689-7/23/\$31.00~\copyright2023 European Union}
\begin{document}
%\ninept
%
\maketitle
\begin{abstract}
As a recent development,  task-oriented dialogues (TODs) have been enriched with chitchat in an effort to make dialogues more diverse and engaging. This enhancement is particularly valuable as TODs are often confined to narrow domains, making the mitigation of repetitive and predictable responses a significant challenge. This paper presents a comparative analysis of three chitchat enhancements, aiming to identify the most effective approach in terms of diversity. Additionally, we quantify the divergence between the added chitchat, the original task-oriented language, and chitchat typically found in chitchat datasets, highlighting the top 20 divergent keywords for each comparison. Our findings drive a discussion on future enhancements for augmenting TODs, emphasizing the importance of grounding dialogues beyond the task to achieve more diverse and natural exchanges.
\end{abstract}
\begin{keywords}
task-oriented dialogue, chitchat, response diversity, entropy, Jensen-Shannon's divergence
\end{keywords}

\section{Introduction}
\label{sec:intro}
Developing dialogue systems that produce diverse responses is a significant and challenging task, particularly when it comes to task-oriented dialogues (TODs). Indeed, due to the specificity of task domains, such dialogues can easily become repetitive: there are only so many ways one can ask a user for their flight destination, for example.  This is why language richness is commonly assessed when comparing dialogue datasets \cite{byrne-etal-2019-taskmaster, DUSEK2020} and evaluating response outputs \cite{nekvinda-dusek-2021-shades, oraby_controlling_2018, jagfeld_sequence--sequence_2018}, using measures such as 
% une métrique est un terme de math un peu spécial cf merriam webster online + mon eù xpérience personnelle en conf. où un matheu m'a fait remarque qu'une métrique c'est qq chose qui a les propriétés d'une distance, tandis qu'une mesure c'est une fonction qui associe un nombre à un objet, sans plus de spécificité, donc ici ne prenons pas de risque parlons de mesure.
the number of unique n-grams, Shannon's text entropy \cite{shannon1948mathematical}, and next-word conditional entropy \cite{manning1999foundations}. To introduce as much natural diversity as possible, human-generated responses are often collected. For instance, the MultiWOZ benchmark \cite{budzianowski-etal-2018-multiwoz} adopts a human-human Wizard-of-Oz style data collection method, while the SGD dataset \cite{rastogi2020SGD} makes use of human annotators to rephrase  dialogues generated based on schemas.

To further diversify TODs, a recent approach has been to enhance them with chitchat \cite{sun2021accentor, young2022fusedchat, chen-etal-2022-ketod}. As an illustrative example, mentioning a few interesting details about the user’s flight destination is likely to yield significant variation, due to the numerous possible destinations, and therefore make responses more engaging.  We note that we consider lexical diversity and engagingness to be positively correlated.  Intuitively, a higher level of lexical diversity leads to less predictable, more interesting responses, which in turn helps users be more engaged. Related research \cite{lexical-div-children1991} has found that children who converse with high lexical diversity are perceived as more appealing, mature and talkative by adults. Furthermore, the addition of lexical diversity has been proposed as a means to improve system responses in chitchat conversations \cite{su-etal-2020-diversifying}, where the primary objective is precisely to maintain user engagement \cite{roller2020open}.

Several approaches currently exist to enhance TODs with chitchat (Section \ref{sub-sec: datasets}), which include incorporating snippets of knowledge-based chitchat, adding complete chitchat exchanges, and including snippets generated by a chatbot trained on a chitchat dataset. It is not immediately clear however which approach is the most effective or which lexical qualities each type of chitchat contributes to TODs, as no cross-comparison has previously been preformed.  This paper aims to bridge this gap by comparing three unique types of chitchat enhancements.

To conduct this study, we utilize Shannon's text entropy and conditional entropy (Section \ref{sub-sec: metrics}) to measure the increased uncertainty (and therefore diversity) found in augmented responses. Additionally, we quantify the divergence between the task language, the added chitchat and typical chitchat, using Jensen-Shannon's divergence.  This allows for a qualitative analysis of the top 20 most divergent tokens for each corpus comparison, shedding light on the most notable lexical contributions of each chitchat type. Finally, based on our findings, we engage in a discussion regarding the next steps to consider when enhancing TODs.

\begin{figure*}[t]
    \centering
    \includegraphics[width=17.8cm]{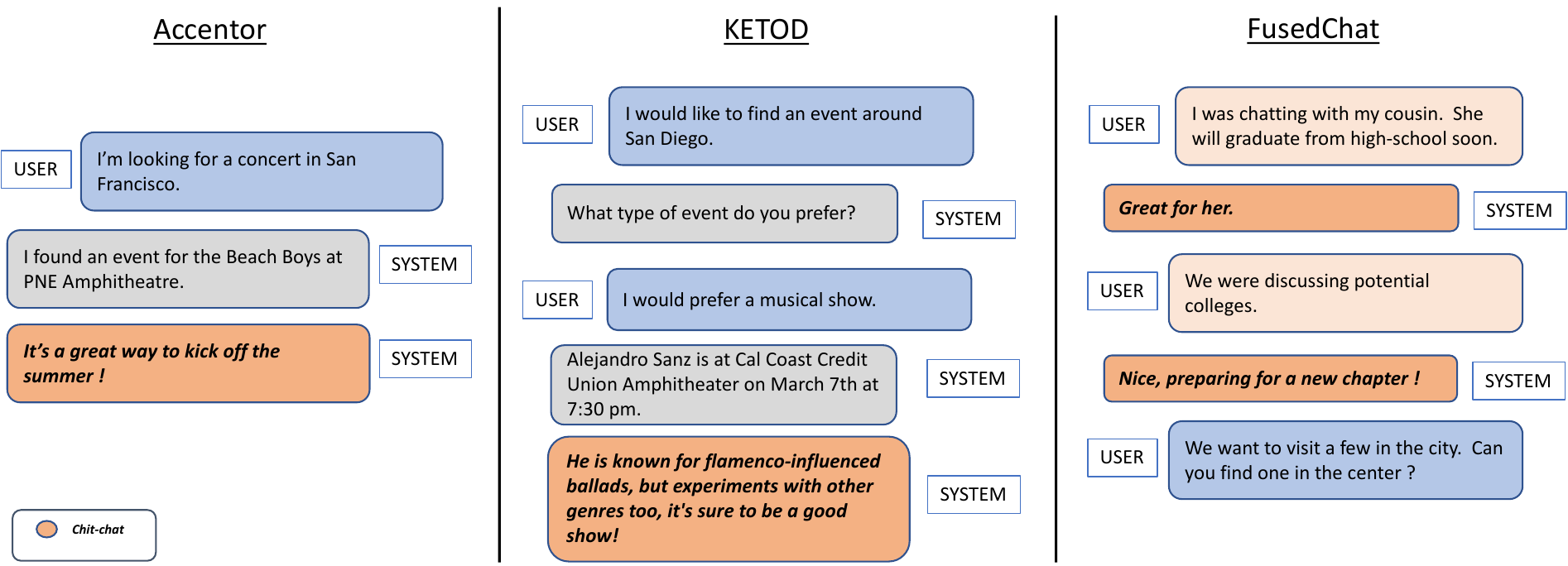}
    \caption{Dialogue examples from each dataset.}
    \label{fig:dialogue-examples}
\end{figure*}

\section{Experimental Setup}
\label{sec:methods}

\subsection{Datasets}
\label{sub-sec: datasets}
Figure \ref{fig:dialogue-examples} showcases an illustrative example of each of the three enhancements assessed in our cross-comparison. We also consider responses from Blended Skill Talk (BST) \cite{smith2020BlendedSkillTalk}, a comprehensive chitchat dataset, as a frame of reference for chitchat responses.

\textbf{Accentor} \cite{sun2021accentor} expands on the SGD dataset \cite{rastogi2020SGD} and comprises 22,825 dialogues.  The authors’ approach is to \begin{bf}automatically generate chitchat candidates additions\end{bf} using a chatbot trained on BST.  These can then be \begin{bf}appended or prepended to the original responses\end{bf}.  To introduce more diversity, the authors automatically filter out frequently occurring candidates and rely on crowd workers to label the remaining candidates as good (ie. social or useful) or bad (ie. inappropriate or misleading). We refer readers to the paper for further details.

Because the chitchat snippets are only \textit{candidates}, we construct a corpus of augmented system responses by randomly selecting a good candidate when several are available and appending (resp. prepending) it to the original task utterance.  If no candidates or only bad candidates are available, the task utterance remains unchanged. To assess the impact of random candidate selection, we repeat the process using 5 different seeds.  We observe minimal impact (1e-4 standard deviation on each metric) and therefore only present the results for one seed.

\textbf{KETOD} \cite{chen-etal-2022-ketod} also extends the SGD dataset. The chosen approach consists in incorporating \begin{bf}chitchat explicitly grounded in Wikipedia into system responses\end{bf}. The methodology involves extracting all entities from each dialogue and employing a retrieval model to fetch the top two Wikipedia articles for each entity. To enrich the system responses, human annotators then select which turns to enhance and incorporate retrieved knowledge snippets,  rephrasing the original response as needed. In cases where annotators find no suitable way to naturally enrich any turns, dialogues are skipped. This process results in a dataset consisting of 5,324 augmented dialogues.

\textbf{FusedChat} \cite{young2022fusedchat} is developed using the well-known MultiWOZ corpus \cite{budzianowski-etal-2018-multiwoz} as its foundation. The chosen approach aims to enhance dialogue diversity by incorporating \begin{bf}chitchat exchanges to introduce or continue pre-existing TODs\end{bf}. This integration results in a reciprocal grounding between TOD and chitchat. The additional chitchat exchanges are created by human annotators:  each annotator assumes the roles of both the system and the user, ensuring a natural flow in the conversation. In some cases, the original task utterances are rephrased to establish a better connection with the chitchat context.  The resulting dataset comprises a total of 10,438 enriched dialogues.

For our analysis, \textbf{BST} \cite{smith2020BlendedSkillTalk} serves as a comprehensive reference chitchat corpus, as it is designed to exemplify multiple qualities within each chitchat conversation. These qualities include being \begin{bf}engaging, knowledgeable, and empathetic\end{bf}. Each conversation is initiated with predefined personas for both participants. Additionally, a pair of utterances is randomly selected from three different chitchat datasets as conversation starters: PersonaChat \cite{zhang-etal-2018-personachat} focuses on maintaining consistent personas throughout the conversations,  Wizard of Wikipedia \cite{dinan2018wizard} draws on expert knowledge sourced from Wikipedia, and EmpatheticDialogues \cite{rashkin-etal-2019-empatheticDialogues} showcases conversations between a Speaker who discusses an emotional situation and a Listener who is tasked with responding in an empathetic manner.

\subsection{Metrics}
\label{sub-sec: metrics}
\textbf{Shannon's text entropy} \cite{shannon1948mathematical} quantifies the average uncertainty of selecting an n-gram from a corpus and has been used to measure lexical diversity in text \cite{shi2022lexical} and in dialogues \cite{DUSEK2020, oraby_controlling_2018}. Compared with simply counting unique n-grams, it also considers their frequencies and distribution, thereby offering a more precise measure of diversity.  When viewing a corpus of responses as a probability distribution over n-grams, a higher entropy is indicative of more uniform distribution, implying greater uncertainties and therefore a higher lexical diversity. A lower Shannon text entropy suggests a more skewed distribution, meaning the corpus contains more frequently repeated n-grams, resulting in less diversity and more predictable responses.

\textbf{Conditional next-word entropy} \cite{manning1999foundations} gives an additional measure of diversity, quantifying the uncertainty of the next token in a sequence given the previous tokens. Typically, if the conditional next-word entropy is high, it implies multiple viable possibilities for  the next token, and therefore more varied and diverse responses. On the other hand, a low conditional next-word entropy suggests a more constrained set of potential next tokens and therefore less diversity.

\textbf{Jensen-Shannon's Divergence} (JSD) \cite{lin1991divergence} is a symmetric version of the Kullback-Leibler divergence \cite{kullback1951information} that evaluates the overlap between two distributions. Based on unigrams, it is commonly used for corpus comparisons \cite{lu-etal-2020-diverging, pechenick2015characterizing},  as it produces divergence scores at the \textbf{corpus} and \textbf{token} levels.  The \textit{k} tokens with the highest divergence scores can be extracted, along with the corpus within which they are most prevalent, allowing us to identify the key divergent words.  Notably, while JSD has been utilized to analyze classroom conversations between teachers and students to assess uptake \cite{demszky-etal-2021-measuring}, its application to comparing collections of dialogue responses remains unexplored.

Experiments are carried out using the lexical-diversity package \footnote{https://pypi.org/project/lexical-diversity}.

\section{Results}
\label{sec: results}

\subsection{Entropy}

\begin{figure}[ht!]
  \centering
  \centerline{\includegraphics[width=8.5cm]{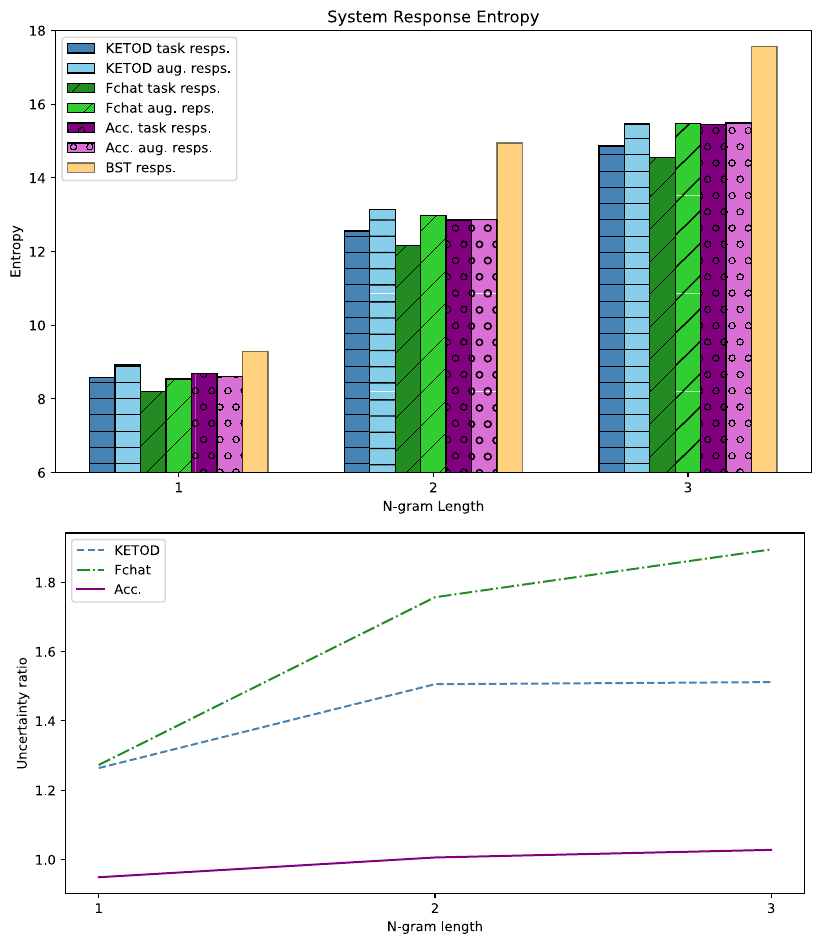}}
\caption{The bar chart presents the entropy for original and augmented responses for our three datasets, and BST. Considering that entropy is a logarithmic measure, the plot below the bar chart shows the uncertainty ratio between the original and augmented responses. For example, when considering trigrams, augmented responses in FusedChat contain approx. 1.89x more uncertainty than their purely task-oriented counterparts.}
\label{fig:entropy}
\end{figure}

Our findings reveal that the introduction of chitchat in KETOD and FusedChat significantly enhances diversity in these datasets, particularly as the n-gram lengths increase (Figure \ref{fig:entropy}). However, despite these improvements, these diversity scores remain considerably lower than those of our reference chitchat responses: the augmented dialogues still exhibit more repetition compared with full-fledged chitchat conversations. This can be attributed to the fact that a limited number of tasks remain the focal point of these dialogues, preventing them from achieving the same level of diversity as the one observed during actual chitchat.

Surprisingly, entropy scores for Accentor (task and augmented responses) show a remarkable similarity.  Augmented responses even show slightly lower entropy in the case of unigram diversity. This unexpected observation signifies that responses containing chitchat exhibit a similar level of diversity (or repetition, depending on perspective) as responses without chitchat. This suggests that the chitchat snippets themselves do not possess significant variation, which could be attributed to the fact they are generated automatically.  Indeed, chitchat systems such as the one used for candidate creation in Accentor tend to output less diverse responses compared with human-created snippets \cite{su-etal-2020-diversifying}. Furthermore, this result puts into perspective human evaluations conducted on this dataset, which indicate that augmented dialogues are perceived as more engaging.  Given the positive correlation we established, we posit that the increase in engagingness is caused by the added chitchat's semantic qualities, rather than its diversity, which we explore in Section \ref{sub-sec: top 20 keywords}, 

To more intuitively grasp the differences in entropy, we plot the uncertainty ratios between the task and augmented responses. For trigrams, the augmented responses in FusedChat exhibit approximately 1.89x more uncertainty than the original task responses, while for Accentor, the augmented responses only show around 1.02x more uncertainty. Accentor's ratio remains relatively unchanged as we increase the n-gram size, while it tends to increase for FusedChat and KETOD. These findings suggest that among the approaches examined, the approach chosen in Accentor contributes little to no extra diversity in system responses, despite these responses being deemed more engaging by human evaluators.

\subsection{Conditional Entropy}

\begin{figure}[ht!]
  \centering
  \centerline{\includegraphics[width=8.5cm]{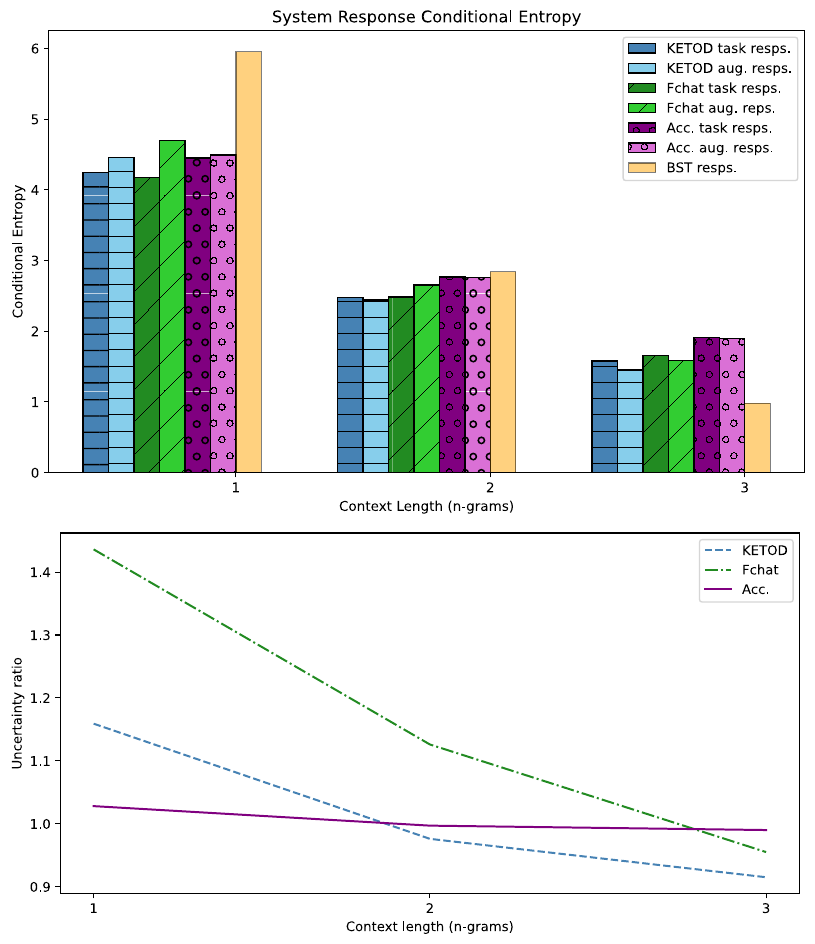}}
\caption{The bar chart presents the conditional entropy for original and enhanced responses, and BST.  The plot below the bar chart should be read as in Figure \ref{fig:entropy}.}
\label{fig:cond_entropy}
\end{figure}

Our findings for conditional entropy (Figure \ref{fig:cond_entropy}) align with our previous results. With a context of a single token, guessing the next token is hardest for augmented responses in FusedChat. The increase in uncertainty is also highest (1.44x). Conversely, Accentor shows the lowest ratio (1.03x), suggesting no change in difficulty for next token prediction.

Furthermore, as the context length is increased, a noticeable trend emerges where collections of responses with higher entropy demonstrate \textit{lower} conditional entropy. This pattern is exemplified by the drastic drops observed in the bars representing BST as the context size increases. This phenomenon can be explained by the fact that as we consider longer n-grams, these typically become more numerous and diverse.  In that case, knowing the preceding \textit{n-1} tokens provides substantial information, thereby increasing the predictability of the next token.

When the context length is set to 2, the uncertainty ratio for KETOD drops below the threshold of 1, indicating \textit{more certainty} in predicting the next token when the responses are augmented.  Considering our previous observation, this suggests the presence of a larger variety of n-grams.

In the case of Accentor, the uncertainty ratios only minimally decrease. Even with a context length of 3, Accentor demonstrates the highest ratio (0.99x), showing that the predictability of the next token remains unaffected and that the added chitchat does not introduce many unique 4-grams. This finding implies that the approach employed by Accentor does not yield a noticeable change in diversity, which aligns with our earlier results.

%Overall, considering the results of entropy and conditional entropy, the FusedChat approach stands out as the one that adds the highest level of diversity. On the other hand, the Accentor approach, despite incorporating chitchat, tends to introduce repetitive elements that do not contribute to increased diversity.

\subsection{Corpus-level JSD}
\label{sub-sec: jsd}

\begin{table}[h]
  \centering
  
  \label{tab:comparison}
  \begin{tabular}{|c|c|c|c|}
    \hline
    & Accentor & FusedChat & Ketod \\
    \hline
    Chitchat vs. Task & 0.217 & 0.323 & 0.393 \\
    Chitchat vs. BST & 0.317 & 0.155 & 0.389 \\
    \hline
  \end{tabular}
  \caption{Divergence scores for each respective dataset between the added chitchat and the original task language, as well as between the added chitchat and typical chitchat found in BST.}
\end{table}

The chitchat in Accentor exhibits the highest similarity to its respective task language, surpassing the other two datasets. In contrast, the chitchat in KETOD demonstrates the highest dissimilarity.  When considering typical chitchat, the chitchat in FusedChat showcases the highest similarity, while the chitchat in KETOD is the most dissimilar, once again.

These findings are intriguing as they indicate that chitchat grounded in an external source of knowledge differs the most from the language commonly observed in both task and chitchat dialogues. While BST also incorporates chitchat grounded in Wikipedia, it likely encompasses different topics. Consequently, KETOD provides the most novel information among the examined datasets.

Moreover, the low divergence observed between the chitchat in FusedChat and the language in BST implies that creating a comparable dataset could involve merging snippets from BST with pre-existing TODS. In this scenario, human annotations would only be required to maintain coherence, eliminating the need for extensive human creative input to generate the complete chitchat exchanges, as is the case in FusedChat.

\subsection{Token-level JSD}
\label{sub-sec: top 20 keywords}

\begin{figure*}[ht!]
\begin{minipage}{17.5cm}
    \centering
    \includegraphics{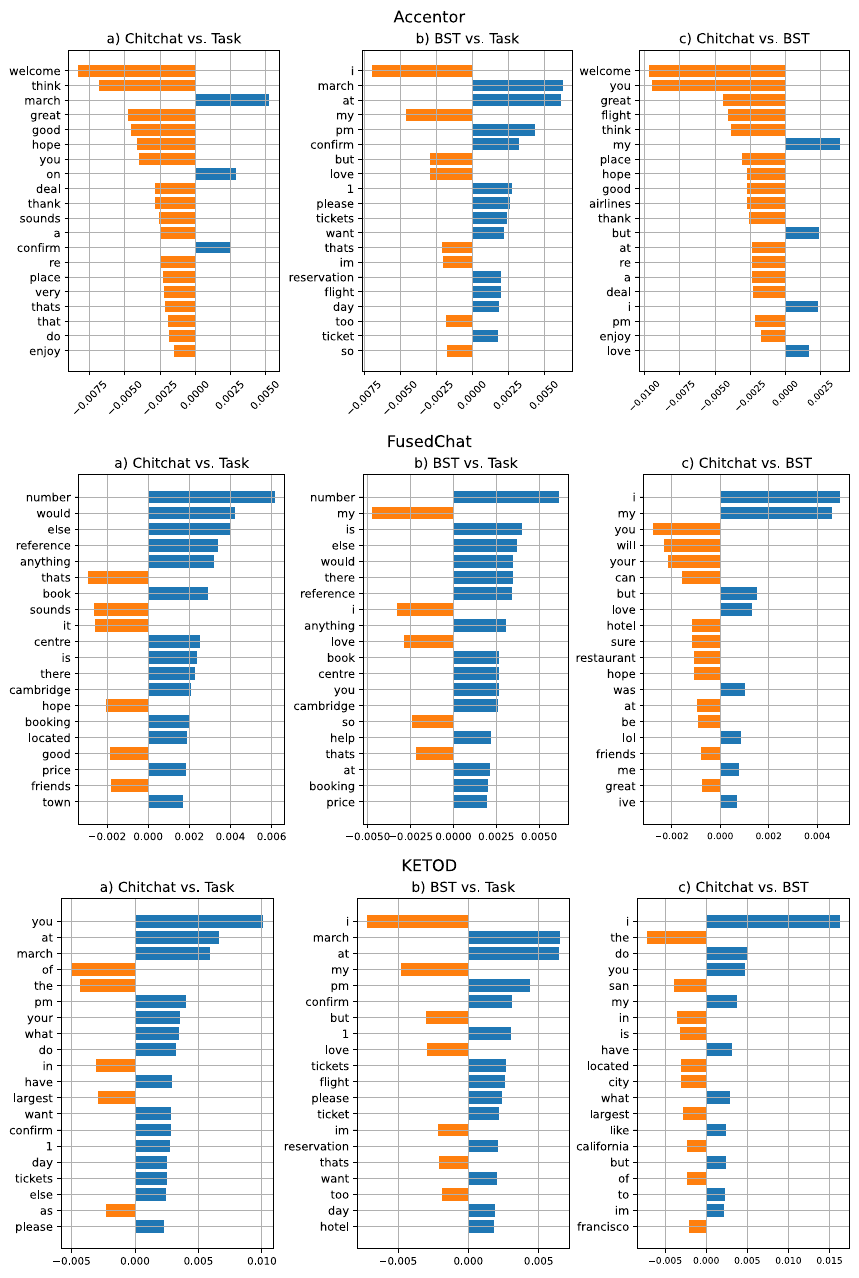} 
\end{minipage}
\caption{Token-level divergences per dataset. The 20 most divergent tokens are shown in each case and are ranked according to their JSD scores. Bar directions are in accordance with each back-to-back chart title.}
\label{fig:Divs} 
\end{figure*}

We identify the top 20 keywords that exhibit the highest levels of divergence in several settings and for each dataset (Figure \ref{fig:Divs}).  With this analysis, we aim to provide valuable insights into the notable semantic differences that characterize each type of chitchat.

In the case of Accentor, the most divergent tokens found in tables a) and c) primarily relate to service-oriented aspects. These tokens include elements from expressions like \textit{you're welcome} and \textit{thank you}, as well as task-specific keywords such as \textit{airlines} and \textit{ticket}. Notably, some of the most divergent tokens convey a positive sentiment, as evidenced by terms like \textit{great}, \textit{good}, and \textit{enjoy}. This semantic quality is what could explain the higher ratings in engagingness given by human evaluators, given the little to no increase in lexical richness. 

In the case of FusedChat, we also observe the presence of positive sentiment in the chitchat responses (chart a), as indicated by words like \textit{fun}, \textit{sounds}, and \textit{good}. Additionally, we notice that the chitchat displays a higher level of responsiveness to user input, utilizing interjections such as \textit{oh} and pronouns such as \textit{thats} (i.e., \textit{that is}) and \textit{it} to refer to previously mentioned information. Upon analyzing chart c), we observe a greater emphasis on the user (\textit{you}, \textit{your}) compared with BST. Moreover, the chitchat appears to be firmly grounded in the MultiWOZ tasks, made evident by references to entities like \textit{hotel}, \textit{restaurant}, and \textit{museum}.

In the case of KETOD, the chitchat primarily focuses on impersonal and factual aspects. In both tables a) and c), we observe the presence of prepositions (such as \textit{of} and \textit{in}), adjectives (like \textit{largest} and \textit{American}), the names of entities (such as \textit{California} and \textit{San Francisco}), and the determiner \textit{the}. These findings indicate that the added chitchat is strongly grounded in task-oriented entities and lacks consideration of the user, contrary to the other approaches.  This demonstrates potential synergy among the different types of chitchat.

Lastly, the analysis of charts b) for each dataset reveals notable patterns in the most divergent tokens within BST. These tokens reflect a stronger inclination towards subjectivity, as evidenced by the presence of words like \textit{I} and \textit{my}.  They also convey a sense of expressiveness with terms such as \textit{love} and \textit{really}, while introducing nuance and argumentativeness through the use of conjunctions like \textit{so} and \textit{but}. These findings shed light on aspects that are lacking in the chitchat used for enhancing task-oriented dialogues, suggesting areas to consider when developing future enhanced task datasets.
   
\section{Discussion}
\label{sec: discuss}
Among the approaches considered, FusedChat emerges as the most effective in achieving diversity: enabling systems to handle both task-oriented and chitchat exchanges proves beneficial for fostering diverse interactions. In contrast, no significant variation in diversity is apparent for Accentor. The higher perceived engagement may in fact be due to the positive sentiment conveyed by the chitchat snippets, rather than response richness. While KETOD may not exhibit the highest level of diversity, its chitchat stands out as the most distinct from both task-oriented and typical chitchat language.  Pushing TODs beyond the constraints of a purely task-oriented database and offering additional grounding therefore offers a valuable enhancement. We propose to expand on this idea.

Indeed, an encouraging approach for enhancing TODs involves adopting a more \textit{situated} dialogue framework that incorporates external knowledge about the world \cite{komeili-etal-2022-internet} and the user. Although the datasets in our comparison integrate chitchat language, they do not fully incorporate the methods used to collect chitchat data, apart from KETOD to a certain extent.  User emotions, a backstory for the interaction, personas that reflect user preferences, and external knowledge could be leveraged to initiate and shape entire TOD conversations, resulting in more diverse and personalized TODs.  We note that this will potentially make modeling TODs with current state-of-the-art approaches \cite{Simpletod_2020, peng-etal-2021-soloist, lin-etal-2020-mintl} more challenging, thereby also driving advancements in TOD system architectures.

This proposed framework is additionally based on the fact that task-oriented and chitchat dialogues are not so distinct when it comes to human communication. In reality, most language is not purely chitchat or task-oriented but a mix of both \cite{brown1983teaching}.  Although the datasets studied move towards a more natural form of communication, a next step would be to further intertwine both modes and modify \textit{both} user and system utterances accordingly, rather than only focusing on the system utterances, as is the case in Accentor and KETOD. 

As a step in this direction, we plan to ground TOD exchanges in plausible situations.  One approach for creating such situations could involve summarizing chitchat from sources like FusedChat: the added chitchat in this dataset is highly task-related and often contains elements of backstory that naturally explain the user's motive for engaging with the system.  However, instead of keeping chitchat and task-oriented exchanges separate, we aim to inject this information directly into the task-oriented user and system turns.  By doing so, we hope to achieve a more diverse and natural dataset of TODs.

\section{Conclusion}
Based on our analysis, we find that FusedChat enhances dialogue diversity the most significantly, while Accentor enhances it the least. Additionally, our examination of the various types of added chitchat reveals some notable qualities in the language added to the datasets, such as positive sentiment, as well as the absence of others, such as nuance and argumentativeness, expressivity, and user consideration in one case. These findings suggest potential synergies between chitchats to look into as future work.

Furthermore, we advocate for the development of more \textit{situated} TODs, grounded in elements commonly found in chitchat datasets: user emotion, user persona, general knowledge, and user backstory. By further intertwining task and chitchat dialogues, we aim to create naturally diverse TOD datasets that are in line with natural human communication.

\bibliographystyle{IEEEbib}
\bibliography{ASRU_arxiv}

\end{document}